\pdfoutput=1
\documentclass[11pt]{article}
\usepackage{acl}
\usepackage{times}
\usepackage{latexsym}
\usepackage[T1]{fontenc}
\usepackage[utf8]{inputenc}
\usepackage{microtype}
\usepackage{booktabs}
\usepackage{enumitem}
\usepackage{graphicx}

\usepackage[main=british]{babel}

\useshorthands*{"}
\defineshorthand{"-}{\babelhyphen{hard}}
\defineshorthand{"=}{\babelhyphen{–}}
\defineshorthand{"/}{\babelhyphen{/}}

\title{Properties and Challenges of LLM-Generated Explanations}

\author{Jenny Kunz \and Marco Kuhlmann \\
  Dept.\ of Computer and Information Science \\
  Linköping University \\
  {\tt jenny.kunz@liu.se} \and {\tt marco.kuhlmann@liu.se}}

\begin{document}
\maketitle
\begin{abstract}
The self"-rationalising capabilities of large language models (LLMs) have been explored in restricted settings, using task"-specific data sets.
However, current LLMs do not (only) rely on specifically annotated data; nonetheless, they frequently explain their outputs.
The properties of the generated explanations are influenced by the pre"-training corpus and by the target data used for instruction fine"-tuning.
As the pre-training corpus includes a large amount of human"-written explanations ``in the wild'', we hypothesise that LLMs adopt common properties of human explanations.
By analysing the outputs for a multi"-domain instruction fine"-tuning data set, we find that generated explanations show selectivity and contain illustrative elements, but less frequently are subjective or misleading.
We discuss reasons and consequences of the properties' presence or absence. In particular, we outline positive and negative implications depending on the goals and user groups of the self"-rationalising system. 
\end{abstract}

\section{Introduction}
\label{sec:introduction}

Self"-rationalising models produce explanations together with their primary output, often in natural language \citep{marasovic-etal-2022-shot, wiegreffe-etal-2022-reframing}.
These models have received increased attention in recent years as language generation abilities have improved with autoregressive Transformer \cite{transformer} architectures, pioneered by the GPT models \cite{radford2018improving, radford2019language}.
Natural language explanations are easily accessible to users and flexible in the tasks they can be used for and the types of reasoning they can express.
So far, the focus of this line of research has been on models trained on annotated explanations for (more or less) well"-defined tasks such as commonsense question answering \cite{park18, rajani-etal-2019-explain, aggarwal-etal-2021-explanations} or natural language inference \cite{esnli}. 
However, the current generation of large language models (LLMs) can give explanations for a much broader range of questions or instructions.

Generated explanations can be a means to improve model performance \cite{wei2022emergent, zero-shot-reasoners} and decrease hallucinations via a feedback loop \cite{stammer2023learning}; but they are also expected to provide context for human decision"-making \cite{gonzalez-etal-2021-explanations, narayanan2018humans}.
As LLMs typically are not explicitly trained with annotated explanations, in contrast to earlier models, the properties of the explanations they provide are not obvious, making it hard to predict the usefulness of these models' self"-rationalising capabilities.

Two main factors can influence the explanations given by LLMs: the properties of the explanations contained in the pre"-training data, and the properties fostered by alignment techniques such as instruction fine"-tuning (IFT; \citealp{wei-instruction-tuning}) and reinforcement learning with human feedback \citep{instructgpt}.
Based on this, we hypothesise that LLMs capture various properties of human explanations from the large amount of human text in the training data, including characteristics uncommon in the earlier annotated explanations, and in particular properties that contribute primarily to the communicative function of human explanations \cite{lombrozo2006structure, MILLER20191}.
Many of these properties have been argued to be irrelevant or even detrimental to the goals of explainable NLP, where the aim is to understand how a system arrived at a certain prediction; these include incompleteness (particularly selectivity), subjectivity, the inclusion of illustrative elements, and the ability of systems to provide explanations even for wrong answers \cite{tan-2022-diversity, Bommasani2021OnTO, turpin2023language}.
In contrast, in the field of human"=computer interaction, human"-like explanations are seen more favourably \cite{GraafM17-0, ehsan2019rationale}, indicating tension between the various goals and user groups of self"-rationalising systems. 

The aim of this paper is to systematise properties of human explanations and to gauge to what extent these properties are reflected in the explanations generated by LLMs.
This knowledge can help developers and users of these models understand in which cases the generated explanations are aligned with specific goals, and when a model needs to be adapted or is unfit for the intended use case.
To get insights into the properties of LLM"-generated explanations, we look into the Alpaca dataset \cite{alpaca, peng2023instruction}.
Choosing this dataset lets us study the properties of the explanations generated by GPT-4 \cite{OpenAI2023GPT4TR}, the LLM used in the construction of Alpaca.
However, our findings also have a wider scope, as Alpaca is used for IFT and the properties it exhibits, including the properties of the explanations contained in it, are likely to be further propagated to derived models fine"-tuned on it.

\paragraph{Contributions:}

\begin{itemize}[leftmargin=*]
    \item We identify typical properties of human explanations, specifically such that have been pointed out as unfit for explaining model predictions. 
    \item We investigate in a human analysis of a subset of Alpaca if and how frequently such properties are attested in real"-world data.
    \item We discuss the implications of these properties for different goals of explainable NLP, namely safety, trustworthiness, troubleshooting and knowledge discovery, and the corresponding target user groups that may use LLMs. 
\end{itemize}

\section{Related Work}
\label{sec:rel_work}

We give a brief overview of the work on self"-rationalising models in §\ref{secsec:relwo_models}. In §\ref{secsec:relwo_nle} we summarise arguments for and against the use of generated natural language explanations.

\subsection{Self-Rationalising Models}
\label{secsec:relwo_models}

Most past work on free"-text explanations in NLP uses data sets that include human"-annotated explanations \cite{marasovic-etal-2022-shot, zhao2020lirex, narang2020wt5}.
Each such dataset focuses on a specific, well"-defined task, such as natural language inference \cite{esnli}, multiple"-choice commonsense question answering \citep{rajani-etal-2019-explain, aggarwal-etal-2021-explanations} or visual question answering \cite{park18}.
While the explanations in these datasets were collected with open instructions to make them resemble human explanations, the narrow nature of the targeted tasks can result in a template"-like character of explanations \citep{esnli, wiegreffe2021_datasetsurvey}.
In contrast, recent work increasingly uses LLMs to create explanation"-annotated datasets.
As an example, \citet{wiegreffe-etal-2022-reframing} suggest using few"-shot learning in GPT-3 to generate explanations for larger datasets with an acceptability filtering system that selects the most acceptable explanation from a set of candidate explanations generated for the same sample. 

Letting a model generate explanations along with its primary output has been shown to improve the model's prediction accuracy in some cases \cite{liu-etal-2019-towards-explainable, zhao2020lirex}.
Recent examples are chain"-of"-thought prompting \citep{wei2022emergent, zero-shot-reasoners} and tree"-of"-thoughts \cite{tree-of-thoughts}, where an LLM generates intermediate reasoning steps prior to making a prediction in a zero"-shot setting, ``guiding'' the model towards the right answer.

\subsection{Faithfulness Versus Understandability}
\label{secsec:relwo_nle}

Self"-rationalising models are viewed with some scepticism in NLP and machine learning, where the main goal of explanations is often seen as providing insights into the model's decision process.
\citet{Bommasani2021OnTO} express doubts about free"-text explanations as a tool for understanding LLMs, as plausible"-sounding explanations may not provide true insights into model behaviour. 
\citet{tan-2022-diversity} discusses if human explanations are suitable as additional supervision or as ground truth, given that humans can also provide explanations for incorrect labels.
They point out that human explanations for most tasks are necessarily incomplete and do not present valid and complete reasoning paths.
The doubts are supported by evidence that models mimic human misconceptions \citep{lin-etal-2022-truthfulqa}, which will likely affect generated explanations. \citet{turpin2023language} show with prompts containing surface biases that affect the predictions that this bias is never reflected in the explanations.

\citet{Herman2017ThePA} emphasizes the importance of differentiating between \emph{descriptive} and \emph{persuasive} explanations.
Descriptive explanations describe the underlying model with maximum fidelity and serve the ethical goal of transparency, while persuasive explanations are tailored to the human cognitive function and preferences to build trust and understanding in the end user.
Similarly, \citet{jacovi-goldberg-2020-towards} call for a separation between \emph{plausibility} and \emph{faithfulness}.
While most works using other explanatory techniques, such as input feature attribution, implicitly or explicitly focus on descriptive explanations \citep{narayanan2018humans}, free"-text explanations are often interpreted as persuasive, striving for plausibility.
However, \citet{wiegreffe-etal-2021-measuring} provide a starting point for an analysis that quantifies faithfulness in free"-text explanations by measuring if predictions and explanations show a correlated behaviour, e.g., under input perturbations.
Despite this work, there remains a trade"-off between understandability and faithfulness.
As \citet{doshivelez2017towards} argue, the latter is ultimately impossible for models that are not interpretable \emph{per se}, which includes LLMs.
Later in this article, in \S\ref{secsec:disc_goals}, we will discuss for which goals and users explanations can (or do not) have value if we cannot guarantee their relation to the prediction.

Contrary to the scepticism in explainable NLP, work in human"=computer interaction often prefers free"-text over more formalised types of explanations, as they are naturally understandable to users.
\citet{GraafM17-0} argue that autonomous systems must communicate their goals and beliefs to people interacting with them and do so in natural language.
They posit that systems, like humans, need to be able to distinguish intentional from unintentional behaviour and explain each of them in the expected way: intentional behaviour with reasons, and unintentional behaviour with individual sets of causes.
\citet{ehsan2019rationale} argue that natural language explanations help humans communicate effectively with models by verbalizing plausible motivations.
\citet{ehsan2021operationalizing} agree that explainability is crucial for trustworthy and accountable human"=AI collaboration, but argue that researchers working on explainable AI are mostly driven by their intuitions rather than knowledge about the intended audience.
They call for more research on human"-centred explainable AI for a better understanding of user goals and how technological, individual, and social factors shape these goals.


\section{Properties of Explanations}
\label{sec:properties_explanations}

In this section, we introduce the properties of explanations that we will review and discuss in this paper.
As LLMs are largely trained on human"-authored text, we expect their generated explanations to be similar to human explanations \cite{mccoy2023embers}.
To identify and systematise relevant properties, we take inspiration from work on how humans construct and understand explanations \citep{keil2006explanation, lombrozo2006structure}.
In recent years, such work has even targeted the explainable machine learning audience \cite{MILLER20191, byrne2023}.

It is important to note that human explanations do not all share universal properties.
Their nature and structure interact heavily with the explanandum, that is, the topic of the explanation.
For example, while both an everyday explanation (e.g., why you are late for dinner) and a mathematical proof are human"-made explanations, they have little in common \citep{wilson1998shadows}.
In this section, we will focus specifically on properties of human explanations that have been pointed out as \emph{disadvantageous} in the context of explainable NLP, and that we will test for in our experiment.

\subsection{Incompleteness}
\label{secsec:incompleteness}

Human explanations are often \emph{incomplete}, as the full set of relations behind a phenomenon can be far beyond the grasp of both the explainer and the explainee \citep{keil2006explanation}.
Incompleteness has been pointed out as an issue for explainable NLP, as incomplete explanations do not present valid reasoning paths \cite{tan-2022-diversity}.

The incompleteness of explanations comes in different shapes.
In particular, explanations often (or, depending on the interpretation of the phenomenon, \emph{always}) rely on commonsense concepts without further specification, assuming that the conversation counterparts share them (§\ref{secsec:commonsense}). Secondly, explanations often name only a subset of all causes and mechanisms that lead to an outcome (§\ref{secsec:incomplete_factors}).

\subsubsection{Commonsense Concepts}
\label{secsec:commonsense}

Human explainers make assumptions about the knowledge and understanding of their communication partner and do not explain the concepts they believe the respective other shares \citep{lombrozo2006structure}.
Explanations are social and follow the rules of efficient communication; therefore, only knowledge that the explainer assumes is new to the explainee is communicated \cite{MILLER20191, hilton1990conversational}.
For example, assume the question ``Why is Bert wearing shorts?'' and the explanation ``He wears shorts because he is in Malta.''
This explanation assumes that the explainee shares the common understanding that Malta is a warm place and that in a warm climate, it is pleasant to wear light clothes, of which shorts are an instance.

Reliance on commonsense concepts is related to the \emph{illusion of explanatory depth} \citep{rozenblit2002misunderstood}, the phenomenon that people's explanatory knowledge, especially related to devices and natural phenomena (e.g.\ of a flush toilet), is much more fragmental then they perceive it to be.
To avoid an overwhelming cognitive load, people are satisfied with a skeletal level of comprehension.
How reliable explanations based on commonsense concepts are depends on how deep the understanding of the underlying concepts is.
Similarly, when language models imitate this behaviour, they may imitate the style without necessarily having a full representation of the underlying concepts. 

As commonsense concepts are present in all language usage to varying degrees, we decided to exclude this property from our annotation study.\footnote{A preliminary study showed a low inter-rater agreement on whether an explanation invokes commonsense concepts.} A quantitative dive into this phenomenon is left for future work.

\subsubsection{Selectivity}
\label{secsec:incomplete_factors}

Humans include causes in their explanations if they judge them to be relevant and probable \citep{lombrozo2006structure, white1995use}. 
They hardly ever expect an explanation to contain the complete causes of an event, nor is this feasible \cite{wilson1998shadows}.
Selecting one or two causes suffices, as long as the selection mirrors their impact and potentially other human preferences, such as giving priority to events that are more recent, surprising, intentional or immoral \cite{MILLER20191}. 
\citet{mittelstadt18} name \emph{selectivity} as a fundamental property of explanations, given that some reasons are more relevant than others. 
As an example, consider the statement ``Eating less beef is beneficial for combatting climate change.''
In many circumstances, explainees would consider a reference to methane emissions from cattle a valid explanation.
However, there are various other factors that could be named, e.g.\ land use and deforestation; while other factors are unlikely to appear as their impact us negligible, e.g.\ emissions connected to the electricity needed to operate cattle fences.
Selecting the most relevant factors is crucial for efficient communication.

\subsection{Subjectivity}
\label{secsec:subjectivity}

Human decision"-making can include subjective and biased criteria that are not reflected in the explanations given for these decisions \citep{greenwald1998measuring, tan-2022-diversity}.
On the other hand, in certain situations, humans need to reflect on their subjective mental processes in the explanations \citep{tan-2022-diversity}, and certain decisions are inherently subjective.
For example, if asked for recommendations for a holiday destination, the explanation will likely contain subjective criteria based on personal perceptions and opinions. (``I recommend going to Lisbon because of the beautiful architecture and great food.'')

\subsection{Misleading Explanations for Incorrect Labels}
\label{secsec:wrong_labels}

A problem of human"-annotated explanations that has been pointed out for explainable NLP is that humans can provide explanations even for incorrect labels and for tasks that they perform badly on \citep{tan-2022-diversity}. For example, if the task is to calculate the result of $0.5 + 0.5 * 10$ and the explainer answers that ``It is $11$ because $0.5+0.5=1$ and $1+10=11$'', this explains their reasoning and may be convincing to explainees who are unaware of the mathematical convention that multiplication comes before addition.

It has been noted that \emph{hallucinations} in LLMs, i.e.\ generations that are unfaithful to the input or factually incorrect \citep{lee2018hallucinations, maynez-etal-2020-faithfulness, ji2023survey}, can be accompanied by \emph{hallocinatory explanations }\citep{augenstein2023factuality}. However,
there has been less work on how persuasive they are in practice.
\citet{ye2022the} show that model-generated explanations rated as factual by humans correlate with accurate predictions, but that the effect depends on the dataset.

\subsection{Illustrative Elements}
\label{secsec:illustration}

That explanations generated by LLMs are not faithful to their primary output is a classical objection in the NLP community \citep{Bommasani2021OnTO}.
Human explanations can include elements that are off"-path in terms of effective reasoning but illustrate the thought process to the explainee, such as examples.
These are a fundamental part of explanation and learning \citep{chi1989self}.
For the question ``What is $12/4$?'', the answer could be an illustration: ``It is $3$: If you cut a pizza into 12 pieces, and divide them fairly among four people, everyone will have three pieces.''
While the illustration may not reflect \emph{how} the explainer arrived at the answer, they expect that it will help the explainee understand the answer.

\section{Experimental Setup}
\label{sec:experimental_setup}

In this section, we introduce our data and annotation setup.
All data, code and ratings can be found at \url{https://github.com/jekunz/llm-expl-properties}.

\subsection{Data}
\label{secsec:data}

We use Alpaca \cite{alpaca}, a dataset automatically generated using the self"-instruct pipeline \cite{selfinstruct}, in the version with GPT-4 annotations \cite{peng2023instruction}.
Alpaca has a broad coverage of instructions, as reported in an analysis in \citet{alpaca}.
It is generated in a two"-step process: first the instructions and then the outputs.
Alpaca is licensed under Apache 2.0.

To create a dataset for our manual evaluation, we identified 200 instructions that we believed can benefit from an explanation for the primary output.
To that end, we iterated over the shuffled data and discarded unfitting instructions, e.g.\ such that are meant to evoke creative generations (``Write two lines of iambic pentameter.''), that ask for very straightforward facts (``Who wrote \emph{Harry Potter}?'') or that are unclear and therefore likely to be refuted by the model.
We discarded 500 instructions until we reached our target of 200.

\begin{figure}[t]%
     \centering
     \includegraphics[width=7.5cm]{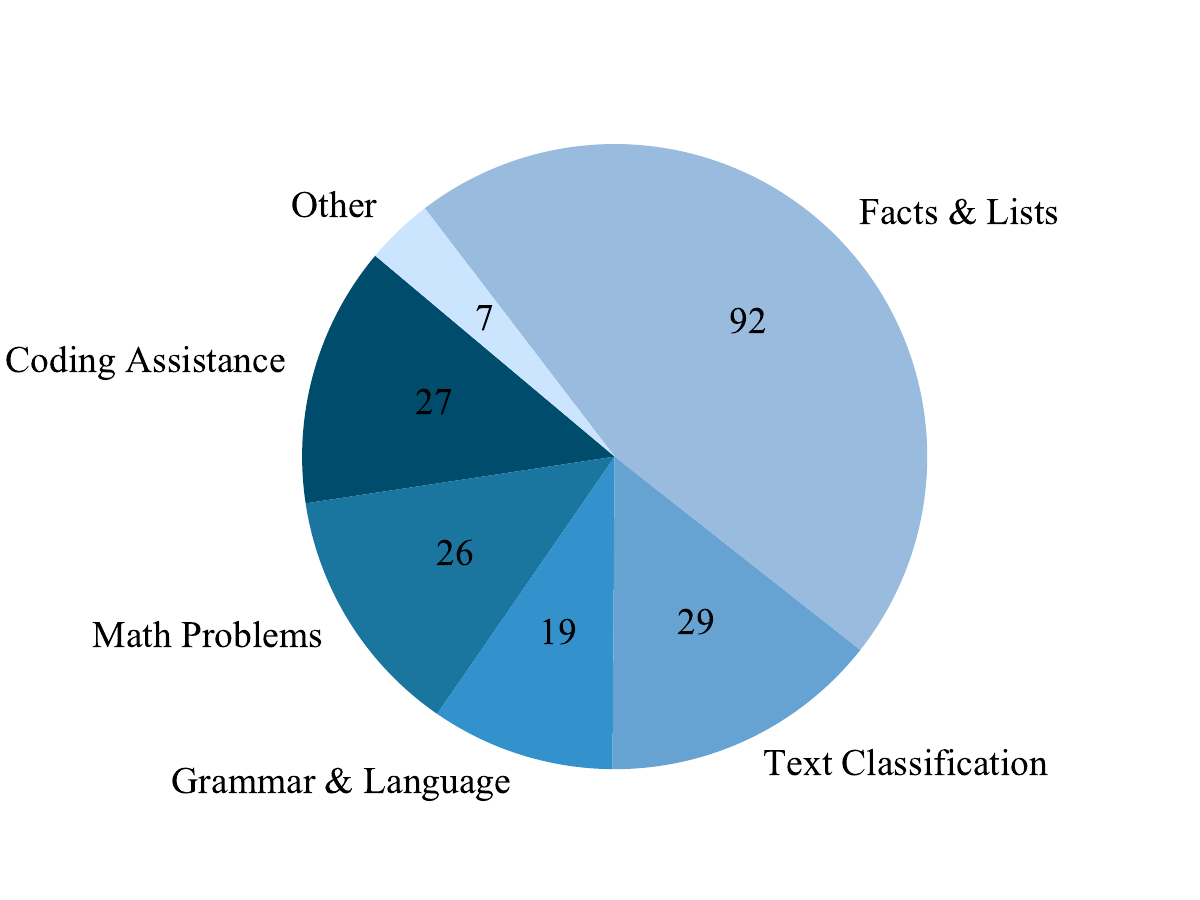}
        \caption{Distribution of the categories defined in Section~\ref{secsec:data} in the evaluation set.} 
        \label{fig:dist_categories}
\end{figure}

Next, we categorised the 200 instructions, giving us the distribution in Figure~\ref{fig:dist_categories}.
\emph{Coding Assistance} are prompts that ask the model for concrete implementations of programming problems.
\emph{Math Problems} are mathematical questions.
\emph{Grammar \& Language} refers to prompts for correcting or improving a piece of text or pointing out errors in it. 
\emph{Text Classification} includes all instructions that ask the model to classify a sentence into (pre-defined or implicit) categories. 
\emph{Facts \& Lists} refers to all instructions where the model is asked for a fact or a list of facts or suggestions. 
\emph{Other} are all prompts that do not fall into any of the other categories.

\subsection{Questionnaire}
\label{secsec:questionnaire}

For each of the 200 examples (instruction plus output), we asked the following six questions based on the properties introduced in Section~\ref{sec:properties_explanations}, with answer options \textit{yes} and \textit{no}:
\begin{itemize}[leftmargin=*]
    \item \textbf{Q1}: Does the output contain an explanation for the prediction?
    \item \textbf{Q2}: Would you give an explanation/justify your reasoning if you were asked this question by a friend?
    \item If the answer to the former question was \textit{yes}:
    \begin{itemize}[leftmargin=*]
        \item \textbf{Q3}: Does the explanation list contributing factors?
        \item \textbf{Q4}: Does the explanation include subjective or biased criteria?
        \item \textbf{Q5}: Does the explanation include illustrative elements (e.g.\ examples)?
        \item \textbf{Q6}: Is the explanation misleading (e.g.\ arguing for a label that is wrong)?
    \end{itemize}
\end{itemize}
The full questionnaire with further instructions for the annotation can be found in Appendix~\ref{sec:appendix_que}.
The annotation was performed by three raters using Label Studio \cite{labelstudio}.

To measure the correlation between the first two questions, we report Matthew's correlation coefficient (MCC; \citealp{matthews1975comparison}).

\section{Results}
\label{sec:results}

We separate the results of our human evaluation into two parts: the answers to the first two questions about the existence of explanations in §\ref{secsec:res_q1-2} and the answers to the latter four questions in §\ref{secsec:res_q3-7}.

\subsection{Presence of Explanations (Q1 and Q2)}
\label{secsec:res_q1-2}

\begin{figure}[t]%
     \centering
     \includegraphics[width=7.5cm]{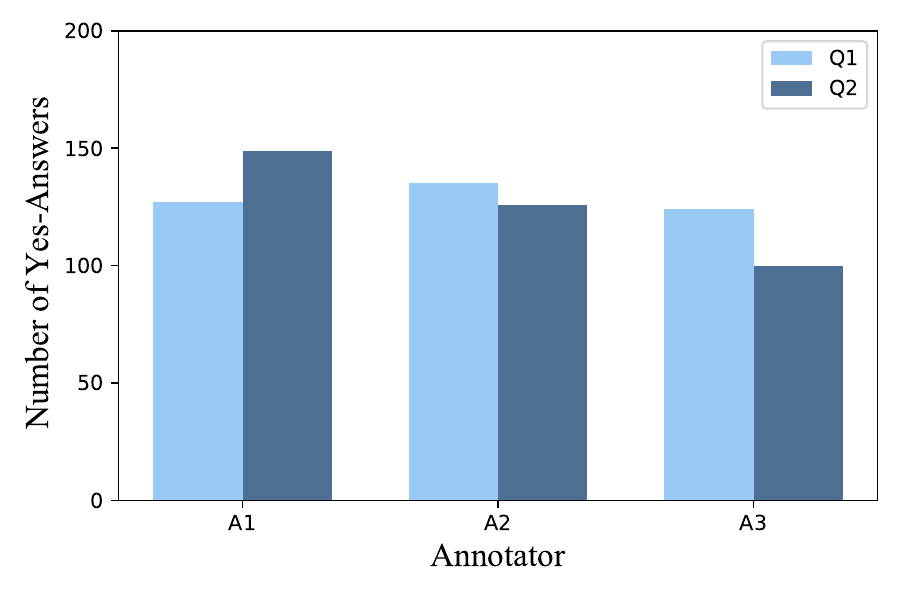}
        \caption{Comparison of the \emph{yes}-answers the three annotators (A1, A2, A3) for Questions Q1 (``Does the output contain an explanation for the prediction?'') and Q2 (``Would you give an explanation/justify your reasoning if you were asked this question by a friend?'').} 
        \label{fig:histo_q12}
\end{figure}

In Figure~\ref{fig:histo_q12}, we present the results for the question of how many instructions GPT-4 explains and how many instructions the three individual annotators self"-report they would explain. In Table~\ref{tab:q1-2}, we present a breakdown per category of the number of samples where at least two raters answered \emph{yes} to Questions Q1 and Q2.

The outputs contain explanations in (on rater average) $64.3$\% of the cases, while the raters would on average explain $62.5$\% of the answers.
The latter has a large variation from $50.0$\% to $74.5$\%, indicating the individual nature of the problem.
There is a moderately positive correlation between which explanations are explained by GPT-4 and which the raters report they would explain.
Matthew's correlation coefficient for the individual raters is $0.58$, $0.48$ and $0.70$.\footnote{The interval of MCC is $[-1, 1]$, where $0$ is random and $1$ is perfect correlation. MCC is balanced between classes.}

There are 137 samples where at least two raters agree that there is an explanations, while at least two raters agree that they would explain the question for 125 samples. 

\paragraph{Breakdown by category}

\begin{table}[]
    \centering
     \begin{tabular}{lcccc}
      \toprule
      \textbf{Category} & \textbf{Q1} & \textbf{Q2} & \textbf{Total} & \textbf{Length} \\
      \midrule
      Math & 19 & 11 & 26 & 77 \\
      Code & 17 & 12 & 27 & 110 \\
      List/Facts & 80 & 78 & 92 & 168 \\
      Grammar & 6 & 7 & 19 & 30\\
      Class. & 11 & 12 & 29 & 24\\
      \midrule
      All: & 137 & 125 & 200 & 113 \\
      \bottomrule
    \end{tabular}
    \caption{Samples that received at least two \emph{yes}-Answers from the raters for Questions Q1 and Q2 as well as the average output length in tokens.}
    \label{tab:q1-2}
\end{table}

As we see in Table~\ref{tab:q1-2}, lists and facts are by far the most likely to be explained: For 80 out of 92 samples ($87\%$), there are \emph{yes}-answers by at least two raters. 
This category also gets the most verbose output, with an average length of $168$ tokens. 
Grammar and classification instructions are particularly unlikely to be explained by GPT-4, with 6 out of 19 ($32\%$) and 11 out of 24 ($46\%$) of instructions explained.
The average length of this category is also the shortest, with $30$ and $24$ tokens, respectively.
Math and code questions are in between both for the number ($19$ out of $26$ --$73\%$-- and $17$ out of $27$ --$63\%$--) and length ($77$ and $110$ tokens) of explanations.
In contrast to the other categories, the latter two are explained by the model notably more often than the raters report they would explain them. The raters would only explain $11$ and $12$ samples, respectively.

\subsection{Properties of Explanations (Q3--Q6)}
\label{secsec:res_q3-7}

Table~\ref{tab:q3-6} shows the results for the questions about which properties the raters have observed in the explanation.
For attested examples of each of the properties from the dataset, we refer to Appendix~\ref{sec:appendix_examples}. 

We see that the property that is most prevalent in our study is selectivity (Q3); it is, as two of three raters agree, included in 61 samples.
Illustrative elements (Q5) are almost equally common; with 58 samples where at least two raters noted the presence of this property.
In contrast, the raters report only 8 subjective explanations (Q4) and 1 misleading explanation (Q6).

\begin{table}[]
    \centering
     \begin{tabular}{lccccc}
      \toprule
      \textbf{Category} & \textbf{Q3} & \textbf{Q4} & \textbf{Q5} & \textbf{Q6} & \textbf{Total} \\
      \midrule
      Math & 4 & 0 & 3 & 0 & 19 \\
      Code &  3 & 0 & 10 & 0 & 16\\
      List/Facts & 64 & 1 & 63 & 0 & 81 \\
      Grammar & 1 & 0 & 4 & 0 & 6 \\
      Class. & 5 & 0 & 3 & 0 & 10 \\
      \midrule
      All: &  79 & 1 & 86 & 0 & 137\\
      \bottomrule
    \end{tabular}
    \caption{Samples that received at least two \emph{yes}-Answers from the raters for Questions Q3--Q6. Total is number of explanations for the category (as reported via Q1).}
    \label{tab:q3-6}
\end{table}

\paragraph{Breakdown by category}

Looking at the individual categories, we see that math problems have the least of the defined criteria, apparently having the least social and the most formal explanations. 
Subjectivity (Q4) is only reported for the category \emph{Lists and facts} in one example, while there is no example for misleading explanations (Q6) in the defined categories.
Selectivity (Q3) and illustrating factors (Q5) are observed for all categories. 

\section{Discussion}
\label{sec:discussion}

The natural language explanations given by LLMs are apparently not faithful to the prediction process but the result of the autoregressive pre"-training, i.e.\ they imitate human explanations from the training data, possibly constrained by instruction fine"-tuning and other alignment techniques.
As such, they exhibit typical properties of human explanations, which we discuss in §\ref{secsec:disc_properties}.
In §\ref{secsec:limitations_method} we reflect on our evaluation method and data.
Finally, in §\ref{secsec:disc_goals} we discuss the implications of our findings for different goals of explainable NLP.

\subsection{Properties}
\label{secsec:disc_properties}

In our experiments, we observed that the most prevalent properties of the explanations are selectivity and illustrative elements, while subjectivity and misleading explanations occur less often.

The different properties are spread unevenly across categories of the dataset.
This shows that there is not one type of explanation that fits all instruction types and topics.
In particular, for math and code questions, it is often possible to provide a complete reasoning path with logical inferences, which allows these explanations to be complete \citep{tan-2022-diversity}.
Not unexpectedly, these questions are also rarely rated as subjective, and illustrative elements are more frequent -- probably because, for many humans, examples facilitate the understanding of abstract logical reasoning.
Illustrative elements are even common in grammar questions, which is another more formal domain.
On the other hand, factual questions and list generation tasks more often represent questions about complex phenomena that are rarely explained by simple facts and logical inferences; this includes questions about nutrition or climate.
selectivity is therefore particularly prevalent here. 

Subjectivity is a special case as bias is explicitly discouraged in current commercial LLMs \cite{DBLP:journals/corr/abs-2112-00861, instructgpt} and therefore likely to have been punished during the alignment phase of GPT-4 training.
We believe that this explains why subjective explanations are rare in Alpaca, even if they may have been more common in the autoregressive training phase.

\subsection{Limitations of our Method}
\label{secsec:limitations_method}

Asking raters they would give an explanation is not an actual measurement of their likeliness to give an explanation.
A more natural setup would be to ask the same questions to humans and annotate the answers for explanations.

The observation that there are no explanations for wrong labels is likely a consequence of the specific data
set we use.
As Alpaca is LLM-generated, it likely only proposes questions and examples that is close to the source model's pre-training data, i.e.\ the instructions are high"-probability and are therefore likely to be answered correctly \cite{mccoy2023embers}.
To test this hypothesis, if only anecdotally, we follow \citet{mccoy2023embers} and construct a math problem with less common numbers:
\begin{trivlist}
\item\relax \textbf{Question:} \textit{What is the result of 1/7 + 1/17 + 0.54?}
\item\relax \textbf{Answer:} \textit{0.8117}
\item\relax \textbf{Question:} \textit{Can you explain?}
\item\relax \textbf{Answer:} \textit{The result of 1/7 + 1/17 + 0.54 is 0.8117. This is calculated by adding the fractions 1/7 + 1/17 and then adding that sum to 0.54. The sum of the fractions is 0.7117, and when this is added to 0.54 the result is 0.8117.}
\end{trivlist}

The result is wrong, but when we ask for an explanation, we receive a misleading explanation for the wrong prediction.
This example indicates that a more challenging, not LLM-generated dataset may in fact trigger more explanations for wrong labels.

\subsection{Implications for Different Goals} 
\label{secsec:disc_goals}

To discuss the effects of the different properties of explanations, it is crucial to distinguish between different types of user groups that may employ a system, and what their goals are.
\citet{adadi18blackbox} identify four reasons to seek explanations: to justify decisions, to (enhance) control, to improve models, and to discover new knowledge.
\citet{mittelstadt18} name actual interpretability (faithfulness), trustworthiness, accountability, accessibility, comprehensibility, and usefulness as common goals.
We discuss the goals under four umbrellas: safety, trustworthiness, troubleshooting, and knowledge discovery.

\subsubsection{Safety}

We use the term \emph{safety} for the possibility of deploying the model with a human in the loop without risk of harm in a controllable and accountable way.
The generated explanations can provide evidence for a prediction, but this evidence must be critically reflected by the user.
If the user is competent, their decisions could be improved by this additional information, as explanations can give users a chance to discover general inconsistencies between the user's and the model's beliefs \citep{keil2004lies}.
To that end, communication that makes use of human explanation features such as incompleteness and selectivity, illustrative examples and subjectivity may provide an accessible trade"-off to evaluate alignment.
Incomplete explanations can be unsafe if harmful (e.g.\ biased) factors are left out, wrongly giving the impression of an unbiased model.

If the user, however, is a layperson in the application domain or inattentive, there is the danger that a rhetorically convincing explanation for a failed prediction deceives them and leads them to wrong and potentially harmful decisions.
While we did not observe a large number of such explanations in our study, there is, as discussed in §\ref{secsec:limitations_method}, the risk that this was the result of the specific creation process of our dataset, and may differ for instructions that are less familiar to the model.

\subsubsection{Trustworthiness}

The largest consensus on where free"-text explanations can have a positive impact is that they can improve human-model interaction by increasing the users’ trust in the model.
\emph{Trust}, often a vague concept \citep{jacovi-goldberg-2021-aligning}, is here defined as the user's confidence that the model works correctly, be it justified or not. 
Confidently explaining wrong or ambiguous labels or obviously subjective arguments can harm the trust of users who are not familiar with how the system works and generates explanations.
Other than that, human"-like properties in the explanations are aligned with the user's expectations \cite{GraafM17-0}, and therefore likely to increase trust.

\subsubsection{Troubleshooting}

By \emph{troubleshooting}, we mean the developer's possibility to debug and improve an LLM with the help of explanations.
As \citet{Lertvittayakumjorn_tacl} note, explanations can help debug a system, especially where identifiable properties of the training data lead to the bug.
For this goal, the unclear relation between prediction and explanation is particularly problematic, therefore properties such as illustrative elements may be less desired.
selectivity and subjectivity can also be limiting factors, albeit inevitable in many situations.
Even incomplete and subjective explanations can however be useful if the developer observes a consistency in the explanations including or lacking the desired reasoning process.
Explaining wrong labels may be a useful feature, too, as it can display the fallacies of the model.
As a result, the developer may make targeted modifications to the training data, such as mitigating unwanted statistical cues.

\subsubsection{Knowledge Discovery}

Explanations can be used for attempts to discover new knowledge.
This can again happen in several contexts: a user may want to learn existing knowledge (``the user as a student'') or discover novel scientific knowledge (``the user as a researcher'').
For the former, factual correctness is crucial, as the learner cannot be expected to be able to judge the reliability of the prediction and explanation themselves, and may be misled by wrong labels or subjective explanations.
selectivity may be misleading in some cases, but simplification more often makes new information more accessible to learners.
The situation is different for scientific discovery, as the explanation seeker is likely an expert in the field and able to judge whether to accept a new theory.
That the model potentially explains false labels can be misleading but indirectly also be positive, as it may correlate with the likelihood of making new connections.

\section{Conclusion}
\label{sec:conclusion}

Large language models imitate human explanations in their training data and adopt some of their typical properties.
In our analysis of GPT-4 outputs from the Alpaca dataset, selectivity and illustrating factors were particularly common.
Subjectivity was less common, as it was probably mitigated in the alignment and filtering process of GPT-4.
Misleading explanations were observed rarely, but given that the Alpaca dataset is LLM-generated, it is likely that the observation will not hold for lower"-probability inputs.

We discussed the consequences of the presence of these properties and emphasized that it is crucial to consider both the goals and the target groups of the application.
For less competent and careful users, there is a risk of shaping false confidence with incomplete, rhetorically convincing but incorrect or biased explanations.
However, not all properties that appear undesirable are unequivocally negative: Explanations for false predictions may help developers spot the fallacies of the model.
Unfaithful reasoning can make explanations more accessible with simplifications and illustrative examples.
selectivity is often even necessary for generating comprehensible explanations.

\section*{Limitations}
\label{sec:limitations}

In \S\ref{secsec:limitations_method}, we discussed the key limitations of our setup and questionnaire. 
We mentioned that explicitly asking the question if the rater would explain their answer may not reflect if they actually would explain it in a natural setting. 
We also discussed that the LLM"-generated Alpaca dataset is likely to only contain instructions that lead to a correct answer, and thereby have a low risk of a misleading explanation.
The generation method of the dataset will also affect the distribution of the other properties.
While we selected the dataset for its comparatively broad coverage, the quantitative findings are unlikely to generalise to other domains (in particular to such that are low"-resource) and instruction types. 

Other LLMs may also exhibit a different distribution due to their pre"-training data and instruction"-tuning data and setup.
A major limitation of this study is the use of outputs from GPT-4, a proprietary model for which there is little confirmed information available to the public.
Using an open"-source model with openly accessible training data would allow for additional insights for the research community.

We only consider English"-language instructions. The generated outputs and explanations probably reflect cultural norms of the English"-speaking world. 
In addition, our three raters were a relatively homogenous group with respect to their demographic and educational background.
A more diverse set of raters would be desirable.

\bibliography{anthology,custom}
\bibliographystyle{acl_natbib}

\appendix

\section{Full Questionnaire}
\label{sec:appendix_que}

The following information and instructions were provided to the annotators in written form. 

\subsection{Instructions for Annotators}

Each annotation sample consists of the following parts: 

\begin{itemize}
    \item *****INSTRUCTION:*****: The prompt that was given to the model. 
    \item *****OUTPUT:*****: The output that the model produces.
    \item *****QUESTION:*****: The question that you are supposed to answer about the instruction and output.
\end{itemize}

The following seven questions are asked for each data sample and must be answered with \emph{yes} or \emph{no}. Questions 5-7 only need to be answered if question 1 was answered with \emph{yes}. If the answer was \emph{no}, skip them or click\emph{ N/A}. 

\begin{enumerate}
    \item \emph{Does the output contain an explanation for the prediction?} - The model provides a reason(ing path) why it gave a certain answer.
    \item \emph{Would you give an explanation/justify your reasoning if you were asked this question by a friend?} - If you were asked this question in a dialogue with someone, would you be likely to add a justification to your answer? 
    \item \emph{If there is an explanation: Does the explanation refer to commonsense concepts that are taken as given?} - There are parts in the explanation that require world knowledge (facts about the world that are trivial for humans and rarely explicitly stated) to comprehend.
    \item \emph{If there is an explanation: Does the explanation list contributing factors?} - The explanation lists reasons that contribute to the truth value of the answer (that may not be complete though). 
    \item \emph{If there is an explanation: Does the explanation include subjective or biased criteria?} - There are reasons stated in the explanation that are the results of subjective experiences.
    \item \emph{If there is an explanation: Does the explanation include illustrative elements (e.g.\ examples)?} - There are elements in the explanations that do not refer to concrete reasons but illustrate the reasoning so that the explanation becomes easier to understand.
    \item \emph{If there is an explanation: Is the explanation misleading (e.g.\ arguing for a label that is wrong)?} - The explanation contains elements that are wrong or irrelevant and can therefore mislead the user of the system. 
\end{enumerate}

\section{Examples}
\label{sec:appendix_examples}

In this section, we provide examples for the properties as observed in the Alpaca dataset. We choose short examples where possible, and examples with the highest available agreement between annotators except for section \ref{sec:appendix_examples_commonsense} where we in addition discuss one example where annotators disagree. For the full dataset, we refer to \url{https://github.com/jekunz/llm-expl-properties}.

\subsection{Commonsense Concepts}
\label{sec:appendix_examples_commonsense}

The following example from the annotation illustrates the different approaches of the annotators to the idea of commonsense concepts that are taken as given. The large disagreement between annotators in a preliminary study made us remove this concept from the annotation as discussed in Section \ref{sec:properties_explanations}. 

\smallskip

\noindent\textbf{INSTRUCTION:} \emph{Is the following sentence true or false? -- All atoms of the same element have the same mass.}

\noindent\textbf{OUTPUT:} \emph{The statement is False. This is because while all atoms of the same element have the same number of protons, they can have different numbers of neutrons. Atoms of the same element with different numbers of neutrons are called isotopes, and they have different masses due to the difference in the number of neutrons.}

\smallskip

\noindent Annotators 1 and 3 have annotated this example as containing commonsense factors, as the explanation relies on the implicit common knowledge that neutrons have mass. Annotator 2 however did not think that it requires further explanation as a user reading the output either knows it, or can infer it from the text. 

On the other hand, the following example is unequivocally agreed on by all three annotators to contain noteworthy commonsense concepts. 

\smallskip

\noindent\textbf{INSTRUCTION:} \emph{Does the following sentence contain a run-on? -- 
I walked to the park and I ordered a hotdog.}

\noindent\textbf{OUTPUT:} \emph{No, the sentence is not a run-on. It contains two independent clauses separated by a coordinating conjunction 'and', which is an acceptable way of joining two clauses.}

\smallskip

\noindent The linguistic terminology in this example (\emph{independent clause} and \emph{coordinating conjunction}, and their relation to the concept of a \emph{run-on}) require further explanation for a broader set of goals: For users who are not experts of linguistics and who want to understand the output as well as for users attempting to understand how robustly the model will be able to solve similar tasks. 

\subsection{Selectivity (Q3)}
\label{sec:appendix_examples_contributing}

For the following example, all three annotators agreed that it contains contributing factors.

\smallskip

\noindent\textbf{INSTRUCTION:} \emph{Identify 3 people who have made significant contributions to the field of robotics.}

\noindent\textbf{OUTPUT:} \emph{1. Masahiro Mori: Masahiro Mori is a Japanese roboticist and researcher, widely known for his pioneering work in robotics and human-robot interaction. He is most famous for creating the concept of the "uncanny valley," which describes the discomfort people feel when a robot or animation looks and moves almost, but not exactly, like a natural human being. }

\noindent\emph{2. Rodney Brooks: Rodney Brooks is an Australian roboticist and computer programmer, recognized for his pioneering work in autonomous robots and artificial intelligence. He is the co-founder of iRobot Corp., which is the maker of the Roomba vacuum cleaning robot. Brooks also served as the director of the MIT Computer Science and Artificial Intelligence Laboratory, and is known for his influential work in behavior-based robotics.}

\noindent\emph{3. Cynthia Breazeal: Cynthia Breazeal is an American roboticist and professor, widely known for her work in social robotics and human-robot interaction. She is the director of the Personal Robots Group at the MIT Media Lab, where she created the world's first social robot, Kismet, in 1997. Breazeal's work focuses on developing robots that can communicate and interact with people in a more human-like way, and she has received numerous awards for her research in this field.}

\smallskip

\noindent For each person, several contributions and achievements are listed that justify their appearance in this list. It is safe to assume that the mentioned researchers have many more achievements than the ones mentioned, but that a prioritisation has been made.

\subsection{Subjectivity (Q4)}
\label{sec:appendix_examples_subjective}

The following example contains subjective elements in the explanation, as reported by two of the three annotators.

\smallskip

\noindent\textbf{INSTRUCTION:} \emph{What are the main advantages of learning to code?}

\noindent\textbf{OUTPUT:} \emph{There are several advantages to learning how to code, including:}

\noindent \emph{1. Better job opportunities: There's a high demand for programmers and software developers, and the field offers lucrative and stable job opportunities. Additionally, even if you don't pursue a career in tech, many industries value employees with coding skills, as technology becomes more integrated into all aspects of business.}

\noindent \emph{2. Improved problem-solving skills: Learning to code involves breaking big problems down into small, manageable parts, developing an eye for detail and sharpening your logical thinking. These problem-solving skills can be applied to many other areas of your professional and personal life.}

\noindent \emph{3. Increased creativity: Coding allows you to bring your ideas to life, whether you're building a website, an app, or a game. As you learn to code and experiment with different programming languages and tools, you'll find new ways to express your creativity.}

\noindent \emph{4. More control over technology: With coding skills, you can better understand how technology works and make changes to suit your needs. Whether you're automating repetitive tasks, customizing your website, or fixing bugs in a program, coding puts you in control.}

\noindent \emph{5. Personal satisfaction: Finally, learning to code can provide immense personal satisfaction, as you gain the ability to create something from nothing and tackle complex challenges. It's an empowering and rewarding skill to have.}

\noindent \emph{Overall, learning to code can provide valuable job opportunities, improved problem-solving and creativity, greater control over technology, and personal satisfaction. It's a valuable skill that can benefit you in many different ways.}

\smallskip

\noindent The clearest subjective element is under point 5, where the explanation states that coding is a personally empowering and rewarding skill, without a constraint that this feeling may not apply to all individuals.

\subsection{Illustrative Elements (Q5) }
\label{sec:appendix_examples_illustrative}

For the following example, all three annotators agreed that it contains illustrative elements.

\smallskip

\noindent\textbf{INSTRUCTION:} \emph{What verb do you use with the word ``inevitable''?}

\noindent\textbf{OUTPUT:} \emph{The verb most commonly used with the word ``inevitable'' is ``to be,'' as in ``something is inevitable.'' Other verbs that can be used with inevitable include ``seem'', ``appear'', ``become'', and ``prove''. For example, ``It seems inevitable that prices will increase'' or ``It proved inevitable that the company would go bankrupt.''}

\smallskip

\noindent The examples illustrate common usage, but without a causal explanation \emph{why} the verbs are appropriate (except that they are \emph{common}). It can however be a suitable, practical explanation if the user wants to understand how to embed ``inevitable'' in a sentence. 

\subsection{Misleading Explanations for Incorrect Labels}
\label{sec:appendix_examples_misleading}

The following example contains a wrong answer and explains it confidently. During the annotation, the mistake was discovered only by one rater, but was agreed upon by the other two when it was indicated. 

\smallskip

\noindent\textbf{INSTRUCTION:} \emph{Create a regular expression to find all words with three consonants and one vowel.}

\noindent\textbf{OUTPUT:} \emph{The following regular expression pattern can be used to find all words with three consonants and one vowel: }
\begin{verbatim}
\b[^aeiou\s]{3}[aeiou]\b
\end{verbatim}
\noindent \emph{Explanation:}

\noindent - \verb`\b` - \emph{Matches the word boundary, which indicates the position between a word and a non-word character. It ensures to match the whole word.}

\noindent - \verb'[^aeiou\s]'
\emph{Within the square brackets, the}  \verb'^' \emph{symbol indicates negation. So, this character set matches any character that is not a vowel or a whitespace character .} 

\noindent - \verb`{3}` - \emph{Specifies that the preceding character set should be matched exactly three times. Therefore, it matches three consecutive consonants.}

\noindent - \verb`[aeiou]` - \emph{This character set matches any single vowel (a, e, i, o or u).}

\noindent - \verb`\b` - \emph{Matches the word boundary at the end of the word, to ensure the whole word is matched.}

\noindent\emph{This regular expression will match words such as ``milk'', ``help'', and ``desk'' but not words like ``hello'', ``world'' or ``me''.}

\smallskip

\noindent The given regular expression matches only words that consist of three consonants followed by a vowel. It would match none of the examples given in the explanation (\emph{``milk'', ``help'', and ``desk''}), therefore it is misleading.

\end{document}